\documentclass[runningheads]{llncs}
\usepackage{graphicx}

\usepackage{tikz}
\usepackage{comment}
\usepackage{amsmath,amssymb} 
\usepackage{color}

\usepackage[accsupp]{axessibility}  

\usepackage{xcolor}

\usepackage[pagebackref,breaklinks,colorlinks]{hyperref}

\begin{document}
\pagestyle{headings}
\mainmatter
\def\ECCVSubNumber{7561}  

\title{TRoVE: Transforming Road Scene Datasets into Photorealistic Virtual Environments} 


\titlerunning{TRoVE}

\author{Shubham Dokania\inst{1}\orcidID{0000-0002-2329-4852} \and
Anbumani Subramanian\inst{1}\orcidID{1111-2222-3333-4444} \and
Manmohan Chandraker\inst{2}\orcidID{0000-0003-4683-2454} \and
C.V. Jawahar\inst{1}}

\author{Shubham Dokania\inst{1} \and
Anbumani Subramanian\inst{1} \and
Manmohan Chandraker\inst{2} \and
C.V. Jawahar\inst{1}}
\authorrunning{S. Dokania et al.}
%
\institute{IIIT Hyderabad, Telangana, India \and
University of California San Diego, CA, USA}
\maketitle

\begin{abstract}
High-quality structured data with rich annotations are critical components in intelligent vehicle systems dealing with road scenes. However, data curation and annotation require intensive investments and yield low-diversity scenarios. The recently growing interest in synthetic data raises questions about the scope of improvement in such systems and the amount of manual work still required to produce high volumes and variations of simulated data. This work proposes a synthetic data generation pipeline that utilizes existing datasets, like nuScenes, to address the difficulties and domain-gaps present in simulated datasets. We show that using annotations and visual cues from existing datasets, we can facilitate automated multi-modal data generation, mimicking real scene properties with high-fidelity, along with mechanisms to diversify samples in a physically meaningful way. We demonstrate improvements in mIoU metrics by presenting qualitative and quantitative experiments with real and synthetic data for semantic segmentation on the Cityscapes and KITTI-STEP datasets. \textit{All relevant code and data is released on github.\footnote{https://github.com/shubham1810/trove\_toolkit}}

\end{abstract}

\section{Introduction}

Computer vision applications, specifically autonomous driving systems, are constantly evolving owing to the rapid progress in the deep learning and machine vision community. At the core of such advancements lies the foundation created by high-quality, structured data, which augments the strengths of sophisticated architectures. While data acquisition and processing require expensive hardware setup and efforts to collect and process, there are several self-driving platforms which provide easier access to large volume of raw and processed data. However, data annotation still poses a huge challenge and is a resource extensive process. Even in the cases when it is feasible, the sheer number of possibilities in real-world diversity makes it unfathomable to observe all variations in object types, scenes, weathers, traffic densities, and sensor configurations. All these possibilities for variations create a near-insurmountable obstacle for dataset curation and annotations for self-driving and road scene scenarios. Use of synthetic data allows creation of such variations and extended diversity for a vast number of scenes, but requires expensive and expert manual efforts for simulations. In this work, we raise the question whether this abundant real-data can be used to automatically create synthetic datasets for training machine learning algorithms and be inclusive of large-variations while preserving real-world structural properties. Mimicking the physical properties of real-data in a synthetic pipeline helps towards minimizing domain gaps, while allowing to generate physically meaningful variations in scenes.

There have been several advancements in the preparation and utilization of synthetic datasets in recent times \cite{shah2018airsim,dosovitskiy2017carla,gaidon2016virtual,ros2016synthia,wrenninge2018synscapes,devaranjan2020meta}. While synthetically generated data may not be a complete substitute for real-world data yet, some works \cite{gaidon2016virtual,varol2017learning,tsai2018learning} discuss the usefulness of augmenting real-world data with synthetic datasets for improved performance . There have been significant improvements in approaches for domain adaptation \cite{tsai2018learning,hoyer2021daformer,chen2021contrastive}, transfer of synthetic-to-real scenes and improvement in photo-realism \cite{roberts2021hypersim,li2020openrooms}. It becomes easier to add different environments, diversity in ambiance, lighting, weathers, and sensors using synthetic data. However, while the data synthesis through simulation is more straightforward than real-world dataset creation, it still requires a significant manual effort. Adding a new scene in such existing pipelines requires expertise to ensure a degree of photo-realism. Recent works highlight learning methods to generate novel trajectories and motion patterns \cite{zheng2020learning,ruiz2018learning,wang2021advsim}. Still, they may not necessarily delineate real-world behavior accurately, especially in complex scenes with varying crowd and traffic densities.

\begin{figure}[t]
\centering
\includegraphics[width=\textwidth]{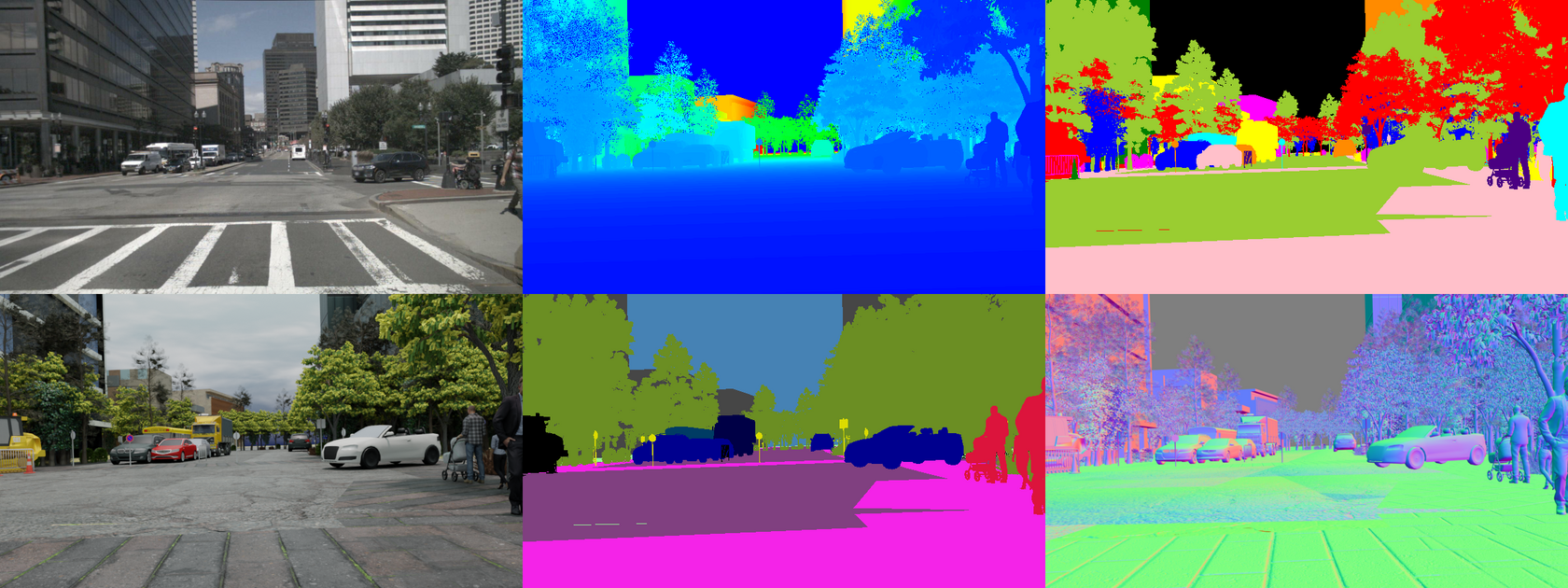}
\caption{A sample of the scene generated with the proposed method. The first image shown is a real-world sample from nuScenes dataset. Using some existing annotation, we are able to generate extensions such as depth, instance segmentation, RGB images, semantic segmentation, and surface normals. These modalities represent a part of the capabilities of the proposed method.}
\label{fig:sample}
\end{figure}

We propose an approach to model synthetic data based on real-world data distributions using available annotations and visual cues, mimicking real-world domain structure and enabling variations in a physically meaningful manner. Unlike most existing works, we show that using information from existing datasets for object placement and behavior can allow for fast construction of virtual environments while preserving the appeal of synthetic data generation systems for efficiency and diversity. We can use the same labels from a real scene to generate a diverse set of annotated data items from each scene (example shown in Figure \ref{fig:sample} and \ref{fig:variation}) with diverse environmental conditions. 

Our approach utilizes the location information from existing real scenes and visual cues available either as annotations or extracted from driving video sequences using currently available approaches in computer vision. We generate high-fidelity environment maps using geographic data available online and supplement this data with extracted cues from existing datasets and scenes. Our physically-based method of scene generation allows us to match different aspects of the scenes such as object positions, orientations, appearances, and ambient factors to recreate virtual environments that mimic real-world. The proposed approach is not restricted to manual design or hand-crafted environments, so it can be extended to virtually any location and complexity configuration for which visual cues can be automatically generated or already available.

\begin{figure}[t]
\centering
\includegraphics[width=0.9\textwidth]{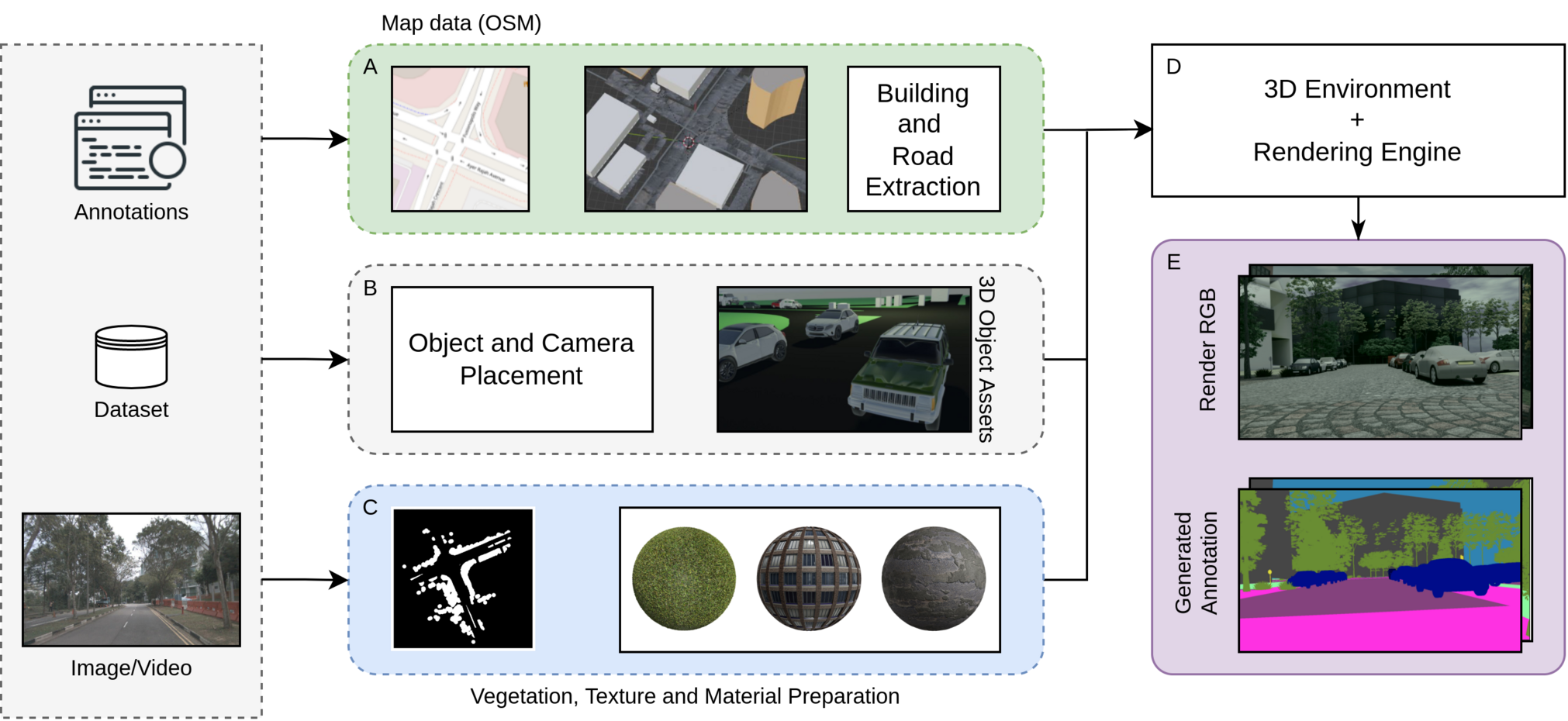}
\caption{An overview of the synthetic data generation pipeline. Given real-world input dataset with annotations for object locations, we follow the process depicted in the above figure. (A) Map data extraction from OSM for building and road extraction. (B) Retrieval of categorical information and 3D object placement in scene with sampled camera poses. (C) Extraction of background data such as vegetation mask in world coordinates and PBR texture preparation. (D) Fusion of artifacts in the 3D environment and initialization of rendering process. (E) Generated RGB data and corresponding annotations stored for further use.}
\label{fig:process}
\end{figure}

A prime example of such a method is to extend existing datasets by utilizing the available annotation from the vast set of high-quality datasets \cite{gahlert2020cityscapes,chang2019argoverse,huang2018apolloscape,8659045} and generate multiple variations for each scene to increase the data volume and annotation modalities available. To support our claims, we outline our approach in the forthcoming sections and use the nuScenes dataset \cite{nuscenes} as a base for the core set of experiments. An overview of our approach is shown in Figure \ref{fig:process}. We use some parts of the already available annotations in nuScenes to generate new environments with traffic and pedestrian behavior similar to that available in the nuScenes dataset while varying the visual information to accommodate a diverse set of configurations. We show qualitative and quantitative analysis over segmentation tasks and compare validation metrics over popular datasets to outline the effectiveness of our data generation strategy for being physically consistent and adhering to real-world distributions. We highlight multiple modalities in our proposed dataset to enable various vision downstream tasks with different sensor configurations.

\section{Related Work}

Data acquisition and annotation for several downstream tasks can be challenging and resource-intensive. Especially for tasks like semantic segmentation, data preparation's cost and time estimate rise rapidly with data volume. There exist many real-world datasets in the community which targets specific problem statements such as vision tasks in indoor scenes \cite{Silberman_ECCV12}, datasets with fine annotations for semantic segmentation and 3D object detection in outdoor scenes \cite{Weber2021NEURIPSDATA,Alhaija2018IJCV,Cordts2016Cityscapes,Cordts2015Cvprw}, and 3D LiDAR point cloud segmentation in urban environments \cite{liao2021kitti,huang2018apolloscape,behley2019semantickitti}. However, considering the cost of expensive annotations, it is often the case that some datasets only focus on specific modalities. For example, the nuScenes dataset \cite{nuscenes} consists of high-quality annotations for object detection, tracking, trajectory prediction, LiDAR segmentation, and panoptic LiDAR segmentation but does not contain fine semantic segmentation annotations due to the sheer volume of available images.

\textit{Simulator-based methods}: Synthetic datasets have been shown to improve performances across a variety of tasks including, but not limited to, object detection \cite{nowruzi2019much,josifovski2018object}, trajectory prediction \cite{zheng2020learning,9197228}, depth estimation \cite{atapour2018real,mayer2018makes}, semantic and instance segmentation \cite{cabon2020virtual,wrenninge2018synscapes}, human pose estimation \cite{varol2017learning}, object 6DoF pose estimation \cite{josifovski2018object}, 3D reconstruction \cite{chang2015shapenet}, tracking and optical flow \cite{wulff2012lessons}. CARLA \cite{dosovitskiy2017carla} is a popular simulator that relies on manually designed environment maps and places 3D object assets for vehicles, pedestrians, and dynamic entities in the environment. CARLA simulates different traffic conditions, variations in lighting, and some weather changes, which are rendered in a photo-realistic manner to provide significant overlap with real-world scenarios. The base version of the CARLA simulator provides a limited number of 3D city environments; different scenarios are simulated, and annotations are generated, which can be either exported to train deep learning models or utilized via their API to evaluate autonomous driving benchmark tasks. LGSVL Simulator \cite{rong2020lgsvl} is a recent addition to the available simulation engines that delivers high-fidelity data for autonomous driving scenarios. LGSVL is built with an integration of Apollo Auto \cite{ap}, which provides various features for interfacing with autonomous driving runtimes. The 3D environment is generated to mimic several real-world locations and integrate multiple sensor types, including RGB, Radar, LiDAR, which can be configured to behave like real-world sensors such as the Velodyne VLP-16 LiDAR. A recent simulation suite built on CARLA is the SUMMIT engine for urban traffic scenarios \cite{9197228}. SUMMIT simulates complex and unregulated behavior in dense traffic environments and utilizes real-world maps to replicate difficult areas like roundabouts, highways, and intersection junctions. SUMMIT uses a context-aware behavior model, Context-GAMMA, an extension of GAMMA \cite{luo2019gamma}, to formulate agents' motion in complex environments for dynamic crowd behavior.

\textit{Non-simulation approaches}: Several works in literature do not rely on simulation of the driving environment directly but provide structured fine annotations. A notable contribution to the community in this area is the Virtual KITTI dataset \cite{gaidon2016virtual} which builds over the popular KITTI dataset \cite{Voigtlaender2019CVPR} and extends the limited amount of annotated information by generating close-to-realistic images for digital-twins of sequences from the KITTI dataset. The Virtual KITTI dataset was recently extended in the Virtual KITTI 2 dataset \cite{cabon2020virtual}, where the quality of images has been improved with a high definition render pipeline and the latest game engine (Unity 2018.4 LTS). The core approach in the Virtual KITTI dataset involves the acquisition of real-world data and measurements from the KITTI MOT benchmarks, then building a synthetic clone of the environments semi-automatically. Then, the objects of interest are placed in the scene manually, and lighting is adjusted to match the real-scene visually. There also exist datasets like SYNTHIA \cite{ros2016synthia}, that do not rely on any visual or geometric cues from real-world datasets and build novel virtual worlds to facilitate synthetic data generation. SYNTHIA provides generated annotations over 13 classes for pixel-level semantic segmentation and frames rendered from multiple view-points in the virtual environment. SYNTHIA dataset consists of large-scale annotations of up to 200k images across four-season settings in the form of video sequences and a random split of data with 13.4k images generated from randomly sampled camera locations across the synthetic map. While the volume and impact of the SYNTHIA dataset is prominent, the data generation process involves exorbitant manual effort. Furthermore, the dynamic entities in the scenes are also programmed manually to capture Spatio-temporal information between different vehicles and pedestrians.

\textit{Learning-based methods}: Some recent approaches focus on imitation training \cite{kishore2021synthetic} for iterative generation of more training data, especially for scenes where the model performs poorly. The work presented in \cite{prakash2021self} deals explicitly with the problem of domain gap in synthetic data by employing self-supervised learning over scene graphs to learn the scene layout and compare generated images with unlabelled images in the target domain. A recurring limitation observed in most datasets tends to be a lack of proper replication of real-world structures in an automated way without learning data or scene-specific layouts. The simulators can construct high-quality environments with variations in multiple factors. They, however, do not replicate the behavior of traffic entities from across a variety of locations around the world, and learned behaviors can only reproduce such motion and trajectory, which is reflective of the training data. In our proposed method for generating synthetic datasets, we leverage visual cues and pre-existing annotations from public datasets and enable the construction of large-scale scenes in virtual environments while mapping driver and pedestrian behavior from actual data into a virtual space.

\section{Our Approach}
\begin{figure}
\centering
\includegraphics[width=0.85\textwidth]{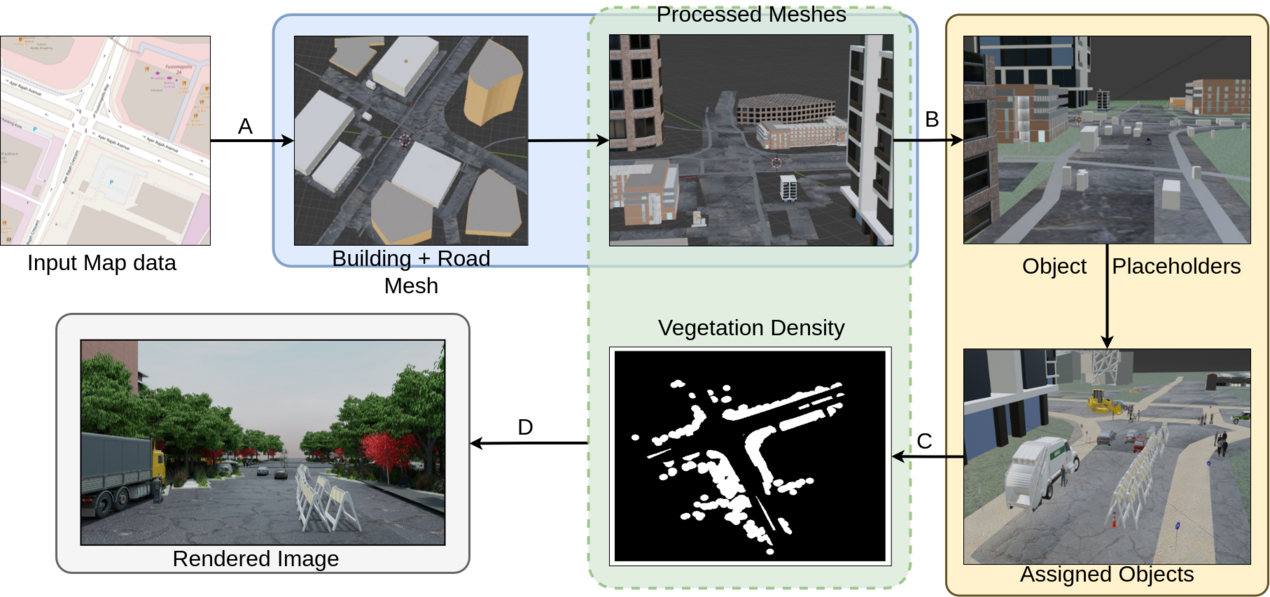}
\caption{Samples from different steps in the proposed approach. Step (A) corresponds to the Section 3.1 for Building and Road processing, (B) corresponds to object placeholder creation and placement, (C) shows vegetation texture density as image, and part of building/road texture in step (A) output. Finally, Stage (D) shows the rendered output for the given scene and the camera pose data.}
\label{fig:step_process}
\end{figure}

We describe the process for generating synthetic data automatically using either existing annotations from public datasets or visual cues from video sequences. We demonstrate the process with an example of the nuScenes dataset \cite{nuscenes}, but the same method can be applied to other datasets with available annotations \cite{chang2019argoverse,Voigtlaender2019CVPR,liao2021kitti}. The core components required to realize the data generation pipeline comprise of the geographic location and objects location/orientation in a given scene or video sequence. It is possible to construct a structured description of the scene and use it as configuration for the data generation process. To ensure diversity in object appearances, we use publicly available free 3D assets for different classes. The process of generating synthetic data using the proposed approach can be broken down into four major parts, which consist of (1) Building the world environment, (2) Placement of objects and camera in the scenes, (3) Applying textures and lighting information, and (4) Rendering and annotation processing. In the following subsections, we describe the steps in detail, referencing the overall process as shown in Figure \ref{fig:process}, and samples from intermediate stages shown in Figure \ref{fig:step_process}. All development of the 3D environment for our dataset is performed in Blender \cite{blender}, an open-source 3D modeling and development software.

\subsection{Building the Virtual Environment}

Assuming the ego-vehicle to be the origin at each scene, we can estimate the geographic location using either GPS data (if available), or offset from scene geometry origin for which GPS information may be available. 

\paragraph{Buildings:} For generating the building placeholder, we refer to data from OpenStreetMaps (OSM) \cite{OpenStreetMap} through a blender add-on (blender-osm \cite{Charles2013}). The meshes are generated without any texture initially. To extend variability in the scene, we choose buildings with an approximately rectangular layout and replace the mesh with a 3D building asset. The 3D assets used in this work were acquired from free-to-use resources on different forums such as \cite{trimbleinc,cgtrader}. To check whether a building (say $b$) qualifies for replacement, we take the points $(x, y)$ from the base plane of $b$ as $\{(x,y) | b_z = 0\}$ and compute the edge length as well as orientations for the building base polygon. Let the edges be denoted by $e_{1,2} = d((x_1, y_1), (x_2, y_2)$ and the orientations be $\theta_{1,2} = \arctan((y_2 - y_1)/(x_2 - x_1))$, where $d$ represents the euclidean distance function. We then compute a histogram of orientations, weighted by the respective edge lengths and select the pairs with close to $\pi/2$ difference. The selected pair with the highest edge weight (lengths) are then chosen to estimate the orientation of the rectangular building base. If no such edges and orientations are found, we assume a complex building outline and mark the building for applying facade texture in a later stage. Optionally, we can also compute the area of the polygon by projecting the base plane on a raster grid and taking ratio of area with the enclosing convex polygon. However, to avoid additional computations, we do not employ this approach in the current pipeline.

\paragraph{Roads:} The road meshes are extracted from OSM map data as well via blender-osm. The road meshes are connected together to form a joint mesh object for the entire road network in the current context. The road width is adjusted to approximately match the road width available from real-world annotations (for nuScenes, extracted from the LiDAR point cloud).

\begin{figure}
\centering
\includegraphics[width=0.85\textwidth]{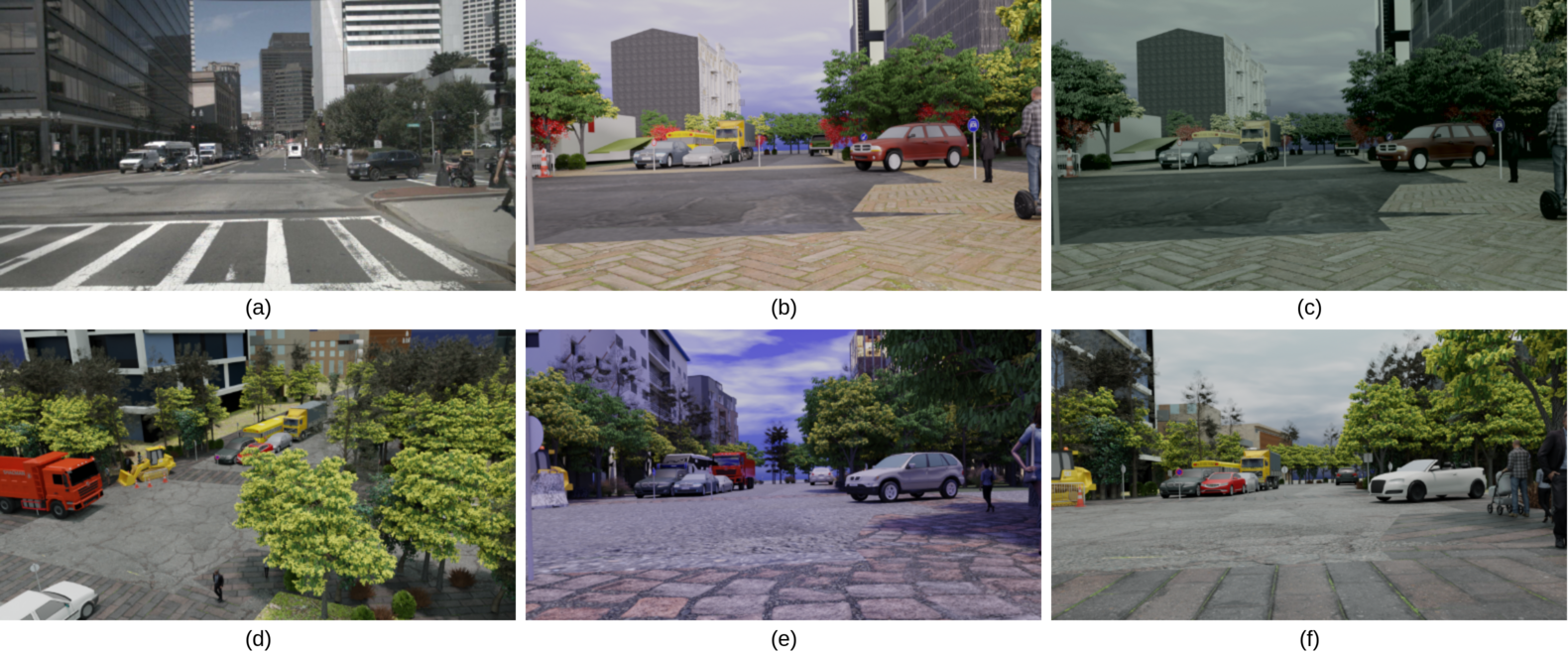}
\caption{Example of variations in synthetic data for the same scene configuration. (a) The real-dataset image depicting a scene from the nuScenes dataset. (b) Sample render with different buildings, vehicles, pedestrians and lighting (c) Rendered image from "b" after applying color transformation for the cityscapes dataset. (d) Render for the same scene from a different camera perspective. (e, f) Additional renders from the scene with variations in vehicles, buildings, lighting etc.}
\label{fig:variation}
\end{figure}

\subsection{Object and Camera Placement}

We utilize the annotations available in the real-world datasets for extracting bounding boxes and camera poses for each scene and lay out the process to replicate the same structure in the virtual environment.

\paragraph{3D Object Placement:} To assign a 3D asset to each bounding box, we estimate a \textit{quality-of-fit} metric based on the Intersection over Union (IoU) for 3D bounding boxes. For a given object bounding box (say $o_i$) and a set of $N$ 3D assets with corresponding bounding boxes (say $\{o_j | j \in 1,...,N\}$) centered at the origin, we scale the asset box such that the largest dimension of the asset box $o_j$ matches that of the query box $o_i$, while preserving aspect ratios. We then compute the 3D IoU metric as follows:
\begin{equation}
    IoU_{3D}(o_i, o_j) = IoU_{xy}(o_i, o_j) * min(z_{o_i}, z_{o_j})
\end{equation}
Where, $IoU_{xy}$ represents the 2D IoU metric for projection on the XY-plane, $z_{o_i}$ is the length of the box along z-axis for object $i$. We are able to use a simplified implementation f the 3D IoU metric due to the known physical properties of both source and target object. We assign the asset with highest $IoU_{3D}$ for a best match, or randomly sample from top-k to induce object diversity.

\paragraph{Camera Pose:} We process the camera matrix and pose information from the source dataset and create virtual clones with similar configuration in the simulated environment. To extend the extent and visual coverage of generated images, we additionally sample camera poses from different vehicles in the scene (along with ego-vehicle), hence diversifying the view-points in a scene.

\subsection{Textures, Lighting and Background}

Texture and Lighting play a critical role towards achieving photo-realism in 3D virtual environment. Additionally, having dense background objects improves the content domain-gap and are a step forward towards realistic distribution of scene geometry.

\paragraph{Textures:} We use high-quality 4K textures and PBR (Physically based rendering) materials from free-to-use forums \cite{polyhaven}. High-resolution maps for color, displacement, roughness, normals, metallic, and emission are available, through which we create BSDF materials for building facades, roads, sidewalks, and ground/terrain. 

\paragraph{Lighting:} High-fidelity materials are important for photo-realism because the pixel-wise distribution of lighting, reflections and color impacts the level of visual perception. Lighting in virtual scenes is very important to accurately model scene dynamics from different view-points. We add different lighting environment models using High Dynamic Range Image (HDRI) to ensure a high range of illumination levels.

\paragraph{Background:} We utilise LiDAR point cloud data from source and construct a 2D location density map for bushes and trees in the scene. To avoid unrealistic occurrences, we remove density data from road area by application of a binary mask. An example of the vegetation density in a scene is shown in Figure \ref{fig:process}(C). In the virtual scene, we apply a probability distribution over the ground plane using the vegetation density map and instantiate trees/bushes. We sample different types and number of trees from the available assets with variations in sizes and sample locations. A similar approach is used for generating traffic signs and poles along the sidewalks in the scene.

\subsection{Data Processing and Training}

Once the preparation of the virtual environment is completed with all objects and entities populated in the scene, we proceed towards rendering and dataset curation stages for synthetic data availability in experiments.

\paragraph{Rendering and Annotations:} We use the Cycles rendering engine in Blender to render the 3D scene and generate multi-modal annotations for semantic and instance segmentation, optical flow, depth estimation, 2D and 3D object detection. We employ the library provided in BlenderProc \cite{denninger2019blenderproc} for annotation extraction. Annotations are generated for 20 classes (including void) namely; sky, car, bus, jeep, truck, van, human, building, road, barrier, ground, cycle rider, construction (vehicle), bushes, trees, motorcycle rider, traffic cone, traffic sign, sidewalk, and void. However, for fair comparison with common benchmarks, we further process the available data into 13 classes as follows: void, car, bus, truck, person, rider, road, sidewalk, building, traffic poles, vegetation, terrain, and sky.

\paragraph{Training Data:} For training and evaluation purpose, we sample a set of 5000 images from our synthetic dataset which have been selected based on the class distribution in annotation maps such that each sample contains a minimum of 6-8 different classes. Additional attention towards imbalance due to background classes is necessary to improve class-wise distribution.

\paragraph{Color Processing:} Since we render images in different lighting conditions, the visual pixel-distribution may vary compared to a dataset we wish to benchmark against. Towards this, we optionally add a color processing stage similar to what was proposed in \cite{reinhard2001color}. We transform the source and target images to L*a*b* color space and adjust the mean and variance of the source domain to a scaled metric between the two domains. A visual example is presented in Figure \ref{fig:variation}(b, c).

\section{Experiments and Results}

For a streamlined analysis and discussion about the experiments, we first outline details about the datasets used in our study and layout the experiment design. Then we analyse the results quantitatively and qualitatively to see how the proposed approach is useful to generate synthetic data that is useful for real-world model training and evaluation.

\subsection{Datasets and Experiments}

For our analysis, we use the Cityscapes \cite{Cordts2016Cityscapes} and KITTI-STEP \cite{Weber2021NEURIPSDATA} datasets. Both of these datasets have been selected for the fairly substantial amount of annotated data available in each. 

\paragraph{Cityscapes:} A widely used dataset for tasks related to visual odometry and perception for road scenes. The dataset provides 5000 finely annotated images for semantic and panoptic segmentation, and an additional 20000 images with coarse annotations with 30 class annotations. Collected over 50 cities in Europe, the dataset provides abundant diversity across scenes with different seasons, and some variations in weather. We use the full training set of the Cityscapes dataset (denoted as $R$) along with a random subset of 1000 images from the training set (denoted as $P$, partial) to show the impact of our synthetic dataset on evaluation of the validation set (denoted as $V$) from the real-world Cityscapes dataset. All images have been re-scaled to 512x256 without loss in aspect ratio for use in training semantic segmentation task. We only use the 5000 finely annotated (2975 for train and 500 for validation) for our experiments.

\paragraph{KITTI-STEP} (Segmenting and Tracking Every Pixel) dataset, is an extension of the KITTI dataset with 21 training and 29 testing sequences from the raw KITTI dataset, based on the KITTI-MOTS \cite{Voigtlaender2019CVPR} dataset and provides semantic as well as panoptic segmentation labels for each image in the sequence along with tracking IDs for non-background objects across frames in a scene. Since this dataset contains more samples compared to the KITTI semantic segmentation benchmark \cite{Alhaija2018IJCV}, we use the 5027 images in the train set and 2981 images in the validation set for our experiments. It is important to note that since the data is extracted from sequences, there is substantial overlap between many frames in the training set which will impact the results, as we shall see in the later analysis. We re-scale the images to a resolution of 620x188 without loss of aspect ratio for our experiments and perform semantic segmentation task. We follow the same notation where the full training data shall be denoted by $R$, partial data of 1000 images as $P$, and the validation set as $V$.

\paragraph{Training Details:} We employ the Deeplab V2 architecture \cite{chen2017deeplab} with the resnet-50 backbone (pretrained with imagenet weights) without CRF. The architecture is kept consistent across all experiments to ensure fairness. For training, 5000 image samples from the generated synthetic dataset are considered, denoted as $S$. Each image is generated at a resolution of 1600x900, then appropriately downscaled and randomly cropped during training to adhere to aspect ratio in the image and match the real data for both training and validation, i.e., 512x256 for Cityscapes and 620x188 for KITTI-STEP. The color transformation scheme mentioned in Section 3.4 is optionally used and will be denoted as $C$ wherever applicable in the results. For combined training of synthetic and real data, we use two methods; we can mix the real and synthetic images in the same batch while training the model (denoted by $M$) or train on synthetic data initially and then fine-tune with the real data (denoted by $F$). For an exhaustive comparison, we present results on all combinations of the settings and report the per-class IoU, mean IoU (mIoU), and global accuracy of each method. For training, we do not employ any additional augmentations apart from randomly cropping the synthetic image to adjust aspect ratio. The models are all trained for 30 epochs, with a batch size of 10 and initial learning rate $1e-04$. The models are trained on a Nvidia RTX 2080Ti GPU using Pytorch-lightning \cite{falcon2019pytorch}. The quantitative results from experiments on Cityscapes and KITTI-STEP are presented in Tables \ref{tab:cityscapes} and \ref{tab:kitti}, respectively.

\setlength{\tabcolsep}{2pt}
\begin{table}
\centering

\caption{Quantitative results for training on real and synthetic data and validation on Cityscapes dataset. We report class-wise IoU, mIoU and global accuracy for the 12 classes (excluding void)}
\resizebox{\linewidth}{!}{%
\label{tab:cityscapes}
\begin{tabular}{|c|c|c|c|c|c|c|c|c|c|c|c|c|c|c|c|} 
\hline
\begin{tabular}[c]{@{}c@{}}\\\\Training \\ Method\end{tabular} & \begin{tabular}[c]{@{}c@{}}Val.\\ Data\end{tabular} & Sky            & Car            & Bus            & Truck          & Person         & Rider          & Road           & Sidewalk       & Building       & \begin{tabular}[c]{@{}c@{}}Traffic \\Poles\end{tabular} & Veget.         & Terrain        & mIoU           & Acc.            \\ 
\hline
R                                                              & V                                                   & 89.25          & 88.86          & 67.66          & 57.16          & 60.15          & 43.66          & 96.51          & 73.88          & 86.83          & 36.77                                                   & 86.49          & 55.71          & 70.25          & 98.88           \\
S                                                              & V                                                   & 27.86          & 41.19          & 3.93           & 4.49           & 23.54          & 14.11          & 71.57          & 18.86          & 61.73          & 1.45                                                    & 70.61          & 28.19          & 30.63          & 95.50           \\
S + R [F]                                                      & V                                                   & 89.03          & 88.86          & 68.27~         & 51.16          & 60.53          & 43.21          & 96.67          & 74.88          & 86.89          & \textbf{39.16}                                          & 86.51          & 57.62          & 70.23          & 98.89           \\
S + R [M]                                                      & V                                                   & \textbf{90.06} & 89.14          & 67.96          & 53.01          & 61.08          & \textbf{44.73} & \textbf{96.84} & \textbf{75.92} & 87.38          & 38.13                                                   & \textbf{87.04} & \textbf{58.49} & 70.82          & 98.93           \\
S + C                                                          & V                                                   & 72.84          & 47.22          & 9.73           & 7.54           & 38.22          & 15.15          & 67.60          & 20.23          & 74.13          & 4.44                                                    & 74.46          & 12.83          & 37.03          & 95.84           \\
S + C + R [F]                                                  & V                                                   & 88.95          & 88.83          & 69.07          & 57.38          & 60.33          & 43.99          & 96.66          & 74.83          & 86.97          & 38.20                                                   & 86.49          & 57.07          & 70.73          & 98.89           \\
S + C + R [M]                                                  & V                                                   & 90.04          & \textbf{89.51} & \textbf{72.09} & \textbf{65.48} & \textbf{61.28} & 41.98          & 96.79          & 75.68          & \textbf{87.46} & 39.08                                                   & 86.99          & 57.35          & \textbf{71.98} & \textbf{98.94}  \\ 
\hline
P                                                              & V                                                   & 86.91          & 86.14          & 29.26          & 32.37          & 55.19          & 33.37          & 95.63          & 68.92          & 84.55          & 30.66                                                   & 84.70          & 49.60          & 61.44          & 98.65           \\
S + P [F]                                                      & V                                                   & 88.00          & 86.71          & 46.71          & 35.02          & 55.22          & 28.40          & 96.03          & 70.62          & 84.99          & 32.80                                                   & 85.03          & 53.16          & 63.56          & 98.70           \\
S + P [M]                                                      & V                                                   & 88.47          & \textbf{87.25} & \textbf{58.44} & \textbf{44.99} & \textbf{57.92} & \textbf{40.24} & \textbf{96.39} & \textbf{73.02} & \textbf{85.86} & 33.80                                                   & \textbf{85.78} & \textbf{54.34} & \textbf{67.21} & \textbf{98.80}  \\
S + C + P [F]                                                  & V                                                   & 87.19          & 86.71          & 49.94          & 36.39          & 55.05          & 26.36          & 95.61          & 68.59          & 85.13          & 33.20                                                   & 85.10          & 48.05          & 63.11          & 98.69           \\
S + C + P [M]                                                  & V                                                   & \textbf{89.05} & 87.05          & 54.07          & 34.73          & 57.24          & 38.33          & 96.32          & 72.52          & 85.84          & \textbf{34.01}                                          & 85.52          & 54.28          & 65.75          & 98.78           \\
\hline
\end{tabular}
}
\end{table}

\subsection{Result Analysis}

\begin{figure}[t]
\centering
\includegraphics[width=0.9\textwidth]{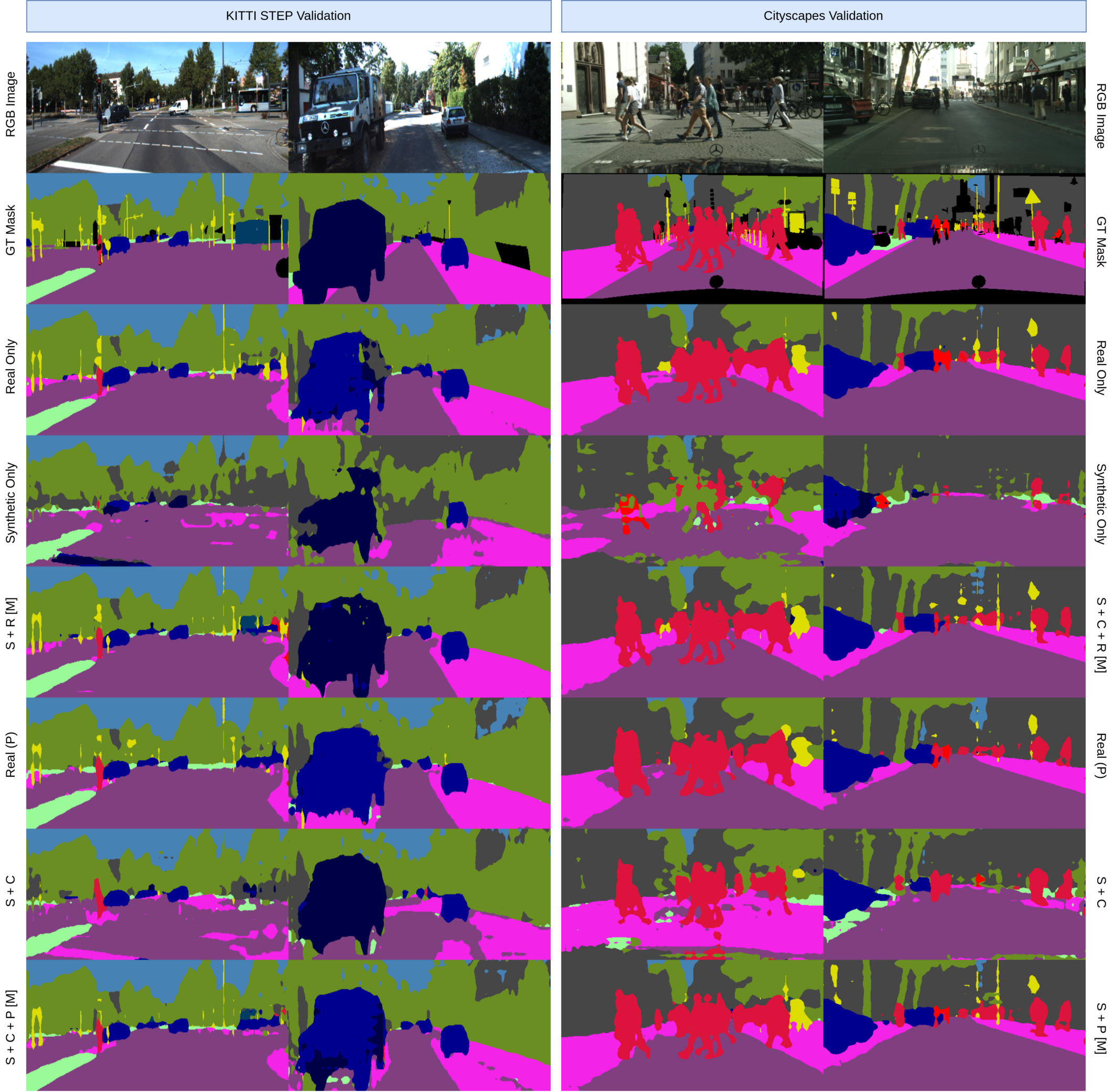}
\caption{Qualitative results for different training strategies using real and synthetic data across KITTI-STEP and Cityscapes datasets. The nomenclature is R: Real, S: Synthetic, P: Real (partial), [F]: Fine-tuned on real, [M]: Real mixed with synthetic, C: Color transformation}
\label{fig:compare}
\end{figure}

In Table \ref{tab:cityscapes} and Table \ref{tab:kitti}, we present the \textbf{mIoU} and IoU per class for the 12 classes (excluding void), along with the global accuracy for each of the different training methods. Furthermore, we present qualitative results on both Cityscapes and KITTI-STEP datasets for some of the methods in Figure \ref{fig:compare}. 

\paragraph{Cityscapes (Full Real):} We notice that training with a mix of real and synthetic data results in a boost of +1.73\% mIoU when the synthetic data has gone through color adjustment. It is interesting to note that while synthetic data in itself may not be enough for achieving high performance (mIoU 30.63\%), whenever combined with real data, helps improve accuracy compared to real data only. This is clear in the qualitative results as well where the model trained with a mix of modified synthetic data and real data is able to generate better segmentation mask for the classes person, sky, traffic pole/sign. It is worth highlighting that for the example in first example (left-sub column) of Cityscapes dataset, the model trained on real data is not able to detect some instance of the person class, while the model trained on the mixture, even with partial, is able to detect the same for this particular sample. When considering autonomous driving scenarios and real-world use cases, this is a crucial detail to consider towards enhancing the performance of deep learning architectures through synthetic data.

\setlength{\tabcolsep}{2pt}
\begin{table}
\centering

\caption{Quantitative results for training on real and synthetic data and validation on KITTI-STEM dataset. We report class-wise IoU, mIoU and global accuracy for the 12 classes (excluding void)}
\resizebox{\linewidth}{!}{%
\label{tab:kitti}
\begin{tabular}{|c|c|c|c|c|c|c|c|c|c|c|c|c|c|c|c|} 
\hline
\begin{tabular}[c]{@{}c@{}}\\Training \\ Method\end{tabular} & \begin{tabular}[c]{@{}c@{}}Val.\\ Data\end{tabular} & Sky            & Car            & Bus            & Truck          & Person         & Rider          & Road           & Sidewalk       & Building       & \begin{tabular}[c]{@{}c@{}}Traffic \\Poles\end{tabular} & Veget.         & Terrain        & mIoU           & Acc.            \\ 
\hline
R                                                            & V                                                   & 91.31          & 86.66          & 0.14           & 18.52          & \textbf{60.99} & 27.93          & \textbf{85.51} & 59.32          & 81.54          & 44.20                                                   & 89.57          & 72.06          & 59.81          & 98.39           \\
S                                                            & V                                                   & 26.44          & 27.29          & 28.85          & 1.95           & 29.94          & 14.34          & 41.21          & 18.96          & 26.73          & 2.53                                                    & 68.75          & 44.81          & 27.64          & 92.96           \\
S + R [F]                                                    & V                                                   & 91.54          & \textbf{87.34} & 1.68           & 13.38          & 60.57          & 27.80          & 84.95          & 59.20          & 82.06          & \textbf{45.49}                                          & \textbf{90.22} & 73.20          & 59.79          & 98.41           \\
S + R [M]                                                    & V                                                   & 91.61          & \textbf{87.34} & \textbf{53.55} & \textbf{28.56} & 60.39          & 30.44          & 85.14          & 58.31          & 81.84          & 42.83                                                   & 90.12          & \textbf{74.26} & \textbf{65.37} & \textbf{98.42}  \\
S + C                                                        & V                                                   & 80.18          & 71.03          & 34.60          & 15.97          & 45.35          & 19.29          & 48.00          & 21.96          & 63.61          & 12.80                                                   & 83.06          & 57.79          & 46.14          & 95.92           \\
S + C + R [F]                                                & V                                                   & 91.41          & 86.99          & 2.23           & 19.98          & 59.58          & \textbf{33.22} & 85.40          & \textbf{59.43} & 81.63          & 45.36                                                   & 90.01          & 73.24          & 60.71          & 98.41           \\
S + C + R [M]                                                & V                                                   & \textbf{91.66} & 85.20          & 45.98          & 25.32          & 59.85          & 29.05          & 84.60          & 57.75          & \textbf{82.24} & 42.99                                                   & 90.03          & 72.55          & 63.93          & 98.38           \\ 
\hline
P                                                            & V                                                   & 90.41          & 84.54          & 0              & 6.16           & 56.09          & 17.61          & 84.39          & 56.23          & 80.23          & 39.54                                                   & 89.62          & 72.30          & 56.44          & 98.28           \\
S + P [F]                                                    & V                                                   & 90.80          & \textbf{85.87} & 12.91          & 8.17           & 56.91          & 16.01          & 83.33          & 54.39          & 80.14          & 42.21                                                   & 89.66          & 72.09          & 57.71          & 98.25           \\
S + P [M]                                                    & V                                                   & 90.64          & 85.31          & 31.85          & 16.13          & 59.17          & \textbf{29.60} & \textbf{84.88} & \textbf{57.21} & 80.90          & \textbf{41.77}                                          & \textbf{89.88} & \textbf{73.25} & 61.72          & \textbf{98.34}  \\
S + C + P [F]                                                & V                                                   & 90.28          & 85.44          & 2.09           & 12.63          & 54.91          & 20.54          & 84.26          & 56.38          & 79.74          & 40.60                                                   & 89.21          & 71.64          & 57.31          & 98.27           \\
S + C + P [M]                                                & V                                                   & \textbf{91.35} & 85.54          & \textbf{48.59} & \textbf{22.65} & \textbf{59.25} & 26.97          & 84.50          & 55.65          & \textbf{81.95} & 39.00                                                   & 89.82          & 71.80          & \textbf{63.09} & 98.33           \\
\hline
\end{tabular}
}
\end{table}

\paragraph{Cityscapes (Partial Real):} We also highlight that for Cityscapes dataset, the model trained with synthetic and partial data achieves a significant improvement (+5.77\% mIoU) over the model trained with just partial data. This result emphasizes on the practical implications of synthetic data where scarcity of real-world annotated data availability may be a bottleneck. Qualitatively as well, the model (S + P [M]) is able to predict sharp segmentation masks for pedestrians crossing the road, compared to the large masks predicted by model (P), trained only on partial real data. The visual results are coherent with the per-class IoU metrics highlighted in Table \ref{tab:cityscapes}. The road, sidewalk, and traffic sign/pole segmentation metric for partial data shows a gap in the performances with and without synthetic data, and the same is observable in example from column 2 (right) in the Cityscapes qualitative sample. The traffic poles are missing almost entirely in the prediction from (P) and the sidewalk structure shows significant noise. Whereas for the same sample, (S + P [M]) shows a higher degree of accuracy and even generates predictions for the poles which have low pixel density.

\paragraph{KITTI-STEP:} To strengthen our claims, we bring attention to the results presented in Table \ref{tab:kitti} where training with synthetic data, assuming full training dataset, shows an improvement of +5.56\% mIoU and with the partial real dataset available only, an improvement of +6.65\% mIoU. The seemingly high difference in the performance for these models can be attributed to the large performance gap in the "bus" and "truck" classes. The reason is that the number of annotated images with bus or truck appearing in the image is very low in the KITTI-STEP dataset. For qualitative confirmation of this case, we highlight column 2 (right) for KITTI-STEP dataset in Figure \ref{fig:compare}. The models trained using only real data misclassify the pixels of the truck object as car, while the model trained with a combination of real and synthetic data accurately segments the object as truck in (S + R[M]). A similar case can be observed for the bus visible in column 1 where only the model trained with synthetic and real combined are able to correctly segment some portion of the bus object. It is important to note that while synthetic data is useful for enhancement of deep learning models, generating such datasets usually requires manual efforts and careful design. However, in this work, we showed a method to generate synthetic dataset fully automatically, hence avoiding the dependency on manual design.

\subsection{Dataset Statistics}

In the experiments presented, we use a sample of 5k images and semantic segmentation annotations, the 5k samples were selected after filtering out ~2.8k samples with low variations in categorical labels per image. However, using the method described in this work, more data can be generated across different modalities including, but not limited to instance and panoptic segmentation, depth estimation, optical flow, surface normal estimation, 2D and 3D object detection, 6DoF object annotations, and 3D object reconstruction. We currently utilise 110 3D assets for different object categories and 40 PBR texture materials for roads, sidewalk, and building facades. In this work, we demonstrate the ability to generate synthetic data based on real-world annotations available in nuScenes dataset, however our approach can be extended to any public dataset to add more diversity in synthetic scenes. For each scene, we sample 20 images and corresponding annotations from different vehicles and camera poses. If OSM map data is available, it takes 25-30s to generate a virtual scene otherwise 30-40s considering an additional API call to OSM server to retrieve map data. Once a virtual scene is generated, it takes 65-85s to render each image of size 1600x900 and the corresponding annotations using a RTX 2080Ti GPU. Essentially, we can generate annotated data in orders of 100,000 within a span of ~1 week with 12 GPUs in parallel.

\section{Conclusions}

We propose a framework for automatic generation of synthetic data for visual perception using existing real-world data. Using a set of 5k synthetically generate images and corresponding semantic segmentation annotations, we show the efficiency of combining synthetic data with real data towards improvements in performance. Given the potential scale of data generation capabilities, various types of data selection strategies can be applied without losing precious annotations. Our approach avoids the pitfalls and limitations of bounded data volumes and variety unlike manually designed virtual environments. Making use of geographical data, our data generation process can be extended to different locations across the globe. Further work in this direction could explore additional modalities and improved photo-realism with more complex scenes generated from a diverse set of datasets and benchmarks on a multitude of tasks. We also highlight another future direction towards better automation by using only visual inputs from a user side.

\section{Acknowledgements}
This work is funded by iHub-data and mobility at IIIT Hyderabad.

%
%
\bibliographystyle{splncs04}
\bibliography{egbib}
\end{document}